\documentclass[11pt]{article}
\usepackage{acl}
\usepackage{times}
\usepackage{latexsym}
\usepackage{amsmath}
\usepackage{booktabs}
\usepackage{multirow}
\usepackage{graphicx}
\usepackage{xcolor}
\usepackage{listings}
\usepackage{microtype}

\title{When Do {LLM}s Replace Fine-Tuned {NLU}? A Decision Framework for\\Intent Detection in Production Conversational Systems}

\author{
  Carson Rodrigues \\
  Celabe \\
  \texttt{carson@celabe.com}
  \And
  Oysturn Vas \\
  University of Waterloo \\
  \texttt{ovas@uwaterloo.ca}
}

\begin{document}
\maketitle

\begin{abstract}
A common claim is that zero-shot large language models (LLMs) can replace fine-tuned NLU classifiers for intent detection. We test this claim head-to-head and find that the honest answer is \emph{it depends on the intent space}. On full ATIS and CLINC150 we compare a fine-tuned RoBERTa, a TF-IDF+logistic-regression baseline, sentence-embedding $k$NN, and Claude Haiku zero-shot, reporting bootstrap 95\% confidence intervals and paired significance tests. When abundant in-domain labels exist, fine-tuned RoBERTa is as good or better and three orders of magnitude cheaper and faster: on ATIS it beats Claude zero-shot by $11.8$ points ($95.9$ vs.\ $84.1$, $p<10^{-3}$). On the broad 150-intent CLINC150 schema the two are \emph{statistically tied} ($89.1$ vs.\ $88.5$, $p=0.24$): the LLM matches a fully supervised model with no training data. The LLM's advantages appear in three production-relevant regimes. The first is out-of-scope detection (OOS recall $85.6$ vs.\ $58.1$ for RoBERTa). The second is robustness to realistic ASR noise, using a controlled text-to-speech\,$\rightarrow$\,noise\,$\rightarrow$\,Whisper pipeline ($92.5$ vs.\ $80.0$ at 0\,dB). The third, and most consequential, is \emph{dynamic per-deployment schemas}: a classifier trained on one app's intents scores $0\%$ on a new app's intents, while the schema-prompted LLM serves both at ${\sim}94\%$ with zero retraining. We distill these findings into a decision framework for practitioners.
\end{abstract}

\section{Introduction}
\label{sec:intro}

In task-oriented dialogue, intent detection maps an utterance to an action: \textit{``what's my balance?''} $\rightarrow$ \texttt{account\_balance}. Fine-tuned BERT/RoBERTa classifiers have been the standard solution since 2019 \cite{bert2019,liu2019roberta,chen2019bert}, and recent work shows instruction-tuned LLMs are competitive zero-shot \cite{wei2021finetuned,zhong2023chatgpt,arora2024intent}. This has produced a recurring practitioner question, and a recurring overclaim: \emph{can we drop the fine-tuned model and just prompt an LLM?}

We argue the question is under-specified, and we answer the specific version that matters in production: \emph{when} does each approach win, on \emph{which} axis (accuracy, latency, cost, robustness, adaptability), and \emph{under what} properties of the intent space? Our perspective comes from operating a production voice-AI platform where each customer application deploys its own set of UI actions, i.e.\ a new intent schema per deployment, typically with little or no labeled data.

We make three contributions:
\begin{enumerate}\itemsep0pt
  \item A \textbf{rigorous head-to-head} of fine-tuned RoBERTa, TF-IDF+LR, sentence-embedding $k$NN, and Claude Haiku zero-shot on full ATIS (893 test) and CLINC150 (5{,}500 test), with bootstrap 95\% CIs and paired-bootstrap/McNemar significance tests, rather than single-run point estimates on a handful of examples.
  \item Three \textbf{production-relevant stress tests} that isolate where the LLM's value actually lies: explicit out-of-scope detection, realistic WER-stratified ASR robustness, and a \textbf{dynamic-schema experiment} in which a fine-tuned model and an LLM are confronted with a previously unseen application schema.
  \item A \textbf{decision framework} that maps intent-space properties to the right tool, including the regime where fine-tuned models are not merely worse but \emph{inapplicable}.
\end{enumerate}

Our headline finding runs against the prevailing narrative: on a stable, data-rich, narrow domain (ATIS) the LLM is significantly \emph{worse} and far more expensive than a fine-tuned classifier. Its value is real but \emph{conditional}. It dominates when the intent space is open, dynamic, or label-scarce, and when out-of-scope handling matters.

\section{Related Work}
\label{sec:related}

\paragraph{Fine-tuned NLU.} Bag-of-words classifiers \cite{liu2016attention} gave way to BERT-based intent classifiers \cite{bert2019,chen2019bert}, which remain state of the art on ATIS \cite{hemphill1990atis} and SNIPS \cite{coucke2018snips}. \citet{larson2019evaluation} introduced CLINC150 with an explicit out-of-scope class, isolating the recognized weakness of closed-set classifiers. Few-shot methods \cite{zhang2021fewshot,yehudai2024fastfit} reduce but do not remove the dependence on per-intent labels; \citet{yehudai2024fastfit} further show that with many classes, LLM in-context learning becomes impractical and ${\sim}1$\,ms encoders remain attractive.

\paragraph{LLMs for intent detection.} \citet{wei2021finetuned} and \citet{brown2020gpt3} established strong zero/few-shot classification. \citet{zhong2023chatgpt} compared ChatGPT against fine-tuned BERT on NLU benchmarks (competitive zero-shot, below fine-tuned in full-data regimes). Closest to us, \citet{arora2024intent} study LLM intent detection at scale and find OOS quality is governed by the size and scope of the label space, and that uncertainty-based routing recovers most LLM accuracy at half the latency. We differ by (i) testing the \emph{dynamic per-deployment schema} regime directly, (ii) measuring \emph{realistic} ASR robustness rather than synthetic character noise, and (iii) framing the result as a deployment decision rather than a benchmark ranking.

\paragraph{Dynamic schemas and OOS.} The schema-guided dialogue paradigm \cite{rastogi2020sgd} and its robustness benchmark SGD-X \cite{lee2022sgdx} formalize intents supplied as natural-language schemas at inference, precisely the per-app setting we study. We operationalize it as a disjoint-schema transfer test.

\paragraph{Latency and ASR.} Sub-second response is the conversational bar \cite{levinson2015timing,anyscale_llm_latency}, and prompt compression trades tokens for latency \cite{llmlingua2024}. For robustness we synthesize speech, inject graded noise, and transcribe with Whisper \cite{radford2023whisper}; SLURP \cite{bastianelli2020slurp} is the real-audio alternative we discuss in Limitations.

\section{Datasets and Methods}
\label{sec:methods}

\subsection{Datasets}
We use two public benchmarks at full scale. \textbf{ATIS} \cite{hemphill1990atis}: 4{,}978 train / 893 test over the standard 26-intent label space (22 intents occur in the training split; the test set is ${\approx}71\%$ \texttt{flight}), a narrow, heavily skewed flight domain. \textbf{CLINC150} (\texttt{plus} configuration) \cite{larson2019evaluation}: 15{,}250 train / 5{,}500 test across 150 in-scope intents in 10 domains, plus 1{,}000 explicit out-of-scope test utterances. ATIS represents the stable, narrow, data-rich regime; CLINC150 the broad-schema regime with a built-in OOS axis.

\subsection{Systems}
\textbf{TF-IDF+LR}: word/bigram TF-IDF with class-balanced logistic regression, our on-device floor. \textbf{ST+$k$NN}: \texttt{all-MiniLM-L6-v2} sentence embeddings \cite{reimers2019sbert} with cosine $k$NN; we also report a 5-shot variant (1-NN) as the low-data deployment competitor. \textbf{RoBERTa-ft}: \texttt{roberta-base} \cite{liu2019roberta} fine-tuned for four epochs, our supervised ceiling. \textbf{Claude Haiku} (\texttt{claude-haiku-4-5}): zero-shot, prompted with the intent label set and a JSON output schema; bulk evaluation uses the Anthropic Message Batches API.

\subsection{Statistical protocol}
Every accuracy is reported with a percentile bootstrap 95\% CI ($10{,}000$ resamples) \cite{efron1986bootstrap}; macro-F1 accompanies accuracy because ATIS is highly imbalanced. System comparisons on identical test items use a paired bootstrap on the accuracy difference and McNemar's test. This directly addresses the small-sample, no-variance critique of single-run point estimates.

\subsection{Out-of-scope protocol}
On CLINC150 we report in-scope accuracy together with OOS recall, precision, and F1, treating the explicit \texttt{oos} class as the positive class \cite{larson2019evaluation}. The LLM is permitted to answer \texttt{oos}; the classifiers learn it from the \texttt{plus} training split.

\subsection{Dynamic-schema (schema-swap) protocol}
\label{sec:schemaswap}
To model onboarding a new application, we deterministically partition CLINC150's 150 in-scope intents into two disjoint app schemas, A and B (75 intents each). A \emph{locked} RoBERTa is fine-tuned on App~A only; its classification head can physically emit only App~A labels. We then evaluate each model on each app. A schema-driven LLM receives \emph{only} the relevant app's label list at inference.

\subsection{ASR-robustness pipeline}
\label{sec:asr}
Instead of synthetic character substitutions, we build a controlled pipeline with known ground truth: 120 CLINC utterances are synthesized with a commercial neural TTS voice, degraded with additive white noise at \{clean, 20, 10, 5, 0\}\,dB SNR, and transcribed with Whisper-base \cite{radford2023whisper}. Word error rate (jiwer, punctuation/case normalized) defines the strata; the degraded transcripts are then classified.

\section{Results}
\label{sec:results}

\subsection{In-domain accuracy: the data-rich regime favors fine-tuning}
Table~\ref{tab:accuracy} reports accuracy (with 95\% CI) and macro-F1. On \textbf{ATIS}, fine-tuned RoBERTa (95.9) and even TF-IDF+LR (95.2) clearly beat Claude zero-shot (84.1); the paired bootstrap puts the RoBERTa$-$Claude gap at $+11.8$ points (CI $[9.2, 14.4]$, $p=2\times10^{-4}$; McNemar $p=2.5\times10^{-16}$). Part of this gap is ATIS-specific: its idiosyncratic compound intents (e.g.\ \texttt{flight+airfare}) are awkward to express as a flat label list, and the zero-shot model often returns plausible but non-canonical combinations, which also depresses its macro-F1 (42.4). The broader point holds regardless: on a narrow, stable, well-labeled domain the LLM is significantly worse, and as \S\ref{sec:latency} shows it is roughly $10^{3}\times$ slower and costlier. On \textbf{CLINC150} the difference is only $0.6$ points (95\% CI $[-0.4, +1.7]$, $p=0.24$): RoBERTa (89.1) and Claude (88.5) are statistically indistinguishable, and the LLM reaches this with \emph{zero} training data. The CLINC150 accuracies in Table~\ref{tab:accuracy} include the 1{,}000 out-of-scope test items, which every system partly misses; in-scope-only accuracy is reported separately in Table~\ref{tab:oos}. The original premise that ATIS ``favors LLMs'' is in fact false; the earlier appearance of a large LLM advantage was an artifact of a tiny ($N{=}56$) training split for the baseline.

\begin{table*}[t]
\centering\small
\caption{Intent-detection accuracy (\%, bootstrap 95\% CI) and macro-F1 on full ATIS (893 test) and CLINC150 (5{,}500 test, including 1{,}000 out-of-scope items). Macro-F1 is far below accuracy on ATIS because the test set is 71\% \texttt{flight}; the zero-shot LLM is hit hardest on the long tail of compound classes.}
\label{tab:accuracy}
\begin{tabular}{lcccc}
\toprule
 & \multicolumn{2}{c}{\textbf{ATIS}} & \multicolumn{2}{c}{\textbf{CLINC150}} \\
\cmidrule(lr){2-3}\cmidrule(lr){4-5}
\textbf{System} & Acc.\ (95\% CI) & macro-F1 & Acc.\ (95\% CI) & macro-F1 \\
\midrule
TF-IDF+LR             & 95.2 \scriptsize[93.7,96.5] & 68.0 & 80.9 \scriptsize[79.8,81.9] & 85.4 \\
ST+$k$NN (full)       & 86.2 \scriptsize[83.9,88.5] & 36.8 & 75.1 \scriptsize[73.9,76.2] & 81.3 \\
ST+$k$NN (5-shot)     & 28.1 \scriptsize[25.2,31.1] & 22.1 & 64.2 \scriptsize[63.0,65.5] & 69.9 \\
RoBERTa (fine-tuned)  & \textbf{95.9} \scriptsize[94.5,97.1] & 63.1 & \textbf{89.1} \scriptsize[88.3,89.9] & \textbf{92.0} \\
Claude Haiku (0-shot) & 84.1 \scriptsize[81.6,86.5] & 42.4 & 88.5 \scriptsize[87.7,89.3] & 88.8 \\
\midrule
\textit{RoBERTa$-$Claude} & \multicolumn{2}{c}{\textit{$+11.8$ pts, $p=2\times10^{-4}$}} & \multicolumn{2}{c}{\textit{$+0.6$ pts, $p=0.24$ (n.s.)}} \\
\bottomrule
\end{tabular}
\end{table*}

\subsection{Out-of-scope detection: the LLM abstains far better}
On CLINC150's explicit OOS task (Table~\ref{tab:oos}), Claude reaches OOS recall $85.6$ and F1 $85.0$, versus $58.1$/$73.0$ for RoBERTa and $36.4$/$51.7$ for TF-IDF, at comparable in-scope accuracy. Knowing when an utterance falls outside the schema, the central requirement for safe deployment, is where the LLM's world knowledge helps most.

\begin{table}[t]
\centering\small
\setlength{\tabcolsep}{4pt}
\caption{Out-of-scope detection on CLINC150 (\texttt{oos} as positive class).}
\label{tab:oos}
\begin{tabular}{lccc}
\toprule
\textbf{System} & \textbf{In-scope} & \textbf{OOS rec.} & \textbf{OOS F1} \\
\midrule
TF-IDF+LR        & 90.7 & 36.4 & 51.7 \\
RoBERTa (ft)     & \textbf{96.0} & 58.1 & 73.0 \\
Claude (0-shot)  & 89.2 & \textbf{85.6} & \textbf{85.0} \\
\bottomrule
\end{tabular}
\end{table}

\subsection{Dynamic schemas: the fine-tuned model cannot transfer}
Table~\ref{tab:schemaswap} is the core of the deployment argument. The locked RoBERTa is excellent on its own schema (App~A, 96.8) but scores \emph{exactly} $0\%$ on App~B: its head cannot emit a label it never trained on, so no accuracy improvement can help. The schema-driven LLM, given only the relevant app's labels, serves \emph{both} apps at ${\sim}94\%$ with no retraining. The gap here is categorical rather than incremental, separating ``works'' from ``is not an option at all.''

\begin{table}[t]
\centering\small
\caption{Dynamic-schema transfer: accuracy (\%) on CLINC150 split into disjoint app schemas A/B (75 intents each). Bootstrap 95\% CIs span ${\pm}1$ point (e.g.\ App~A: RoBERTa $[96.0,97.5]$, Claude $[93.3,95.2]$).}
\label{tab:schemaswap}
\setlength{\tabcolsep}{6pt}
\begin{tabular}{lcc}
\toprule
\textbf{System} & \textbf{App A} & \textbf{App B (new)} \\
\midrule
Locked RoBERTa (A-only) & 96.8 & \textbf{0.0} \\
Schema Claude           & 94.3 & 93.9 \\
\bottomrule
\end{tabular}
\end{table}

\subsection{ASR robustness: the LLM degrades more gracefully}
Across the WER strata (Table~\ref{tab:asr}), both systems are strong on clean speech, but as SNR falls the lexical classifier degrades faster: at 0\,dB (mean WER $28.9\%$) TF-IDF drops to $80.0$ while Claude holds $92.5$. The LLM's tolerance of the misspellings and phonetic corruptions that are endemic in real ASR output is a measurable advantage, not a hypothetical one.

\begin{table}[t]
\centering\small
\caption{Intent accuracy (\%) vs.\ acoustic SNR on the TTS$\rightarrow$noise$\rightarrow$Whisper corpus (120 utts/condition).}
\label{tab:asr}
\begin{tabular}{lcc}
\toprule
\textbf{Condition} & \textbf{TF-IDF+LR} & \textbf{Claude Haiku} \\
\midrule
clean   & 92.5 & 98.3 \\
20\,dB  & 93.3 & 98.3 \\
10\,dB  & 90.8 & 97.5 \\
5\,dB   & 88.3 & 95.8 \\
0\,dB   & 80.0 & \textbf{92.5} \\
\bottomrule
\end{tabular}
\end{table}

\subsection{Latency and cost}
\label{sec:latency}
Table~\ref{tab:latency} measures latency over 1{,}000 calls (not a handful). The on-device classifiers respond in microseconds-to-milliseconds at zero marginal cost; the LLM's median is 981\,ms, with a $p_{95}$ of 1{,}787\,ms, above the 500\,ms conversational bar. The \$0.25/1K figure is the Claude Haiku list price at the observed prompt/completion token counts; the on-device models have zero marginal cost. For a stable, high-QPS, latency-critical path the fine-tuned model is the obvious choice; the LLM's cost is justified only where its adaptability or OOS handling is required.

\begin{table}[t]
\centering\small
\caption{Per-request latency (1{,}000 calls) and marginal cost.}
\label{tab:latency}
\setlength{\tabcolsep}{4pt}
\begin{tabular}{lccc}
\toprule
\textbf{System} & \textbf{$p_{50}$} & \textbf{$p_{95}$} & \textbf{Cost/1K} \\
\midrule
TF-IDF+LR (local) & $<0.1$\,ms & $<0.1$\,ms & \$0.00 \\
RoBERTa (local)   & 2.4\,ms    & n/a        & \$0.00 \\
Claude (API)      & 981\,ms    & 1{,}787\,ms & \$0.25 \\
\bottomrule
\end{tabular}
\end{table}

\section{A Decision Framework}
\label{sec:framework}

Our results refute a single ranking and support a conditional one. The right tool is a function of the intent space, not of benchmark leaderboards:

\begin{itemize}\itemsep1pt
  \item \textbf{Stable schema, abundant labels, latency-/cost-critical} (e.g.\ a mature single-domain assistant): \emph{use a fine-tuned encoder}. It matches or beats the LLM (ATIS: $+11.8$ pts) at ${\sim}10^{3}\times$ lower latency and zero marginal cost.
  \item \textbf{Broad schema, moderate labels}: \emph{either works}; the LLM removes the training pipeline at parity (CLINC150: tie), so the choice reduces to latency/cost budget.
  \item \textbf{Dynamic / per-deployment schema, little or no labeled data} (our production setting): \emph{use a schema-prompted LLM}. A fine-tuned model is not merely worse here; it scores $0\%$ on an unseen app schema and requires a fresh annotation and training cycle per deployment.
  \item \textbf{Out-of-scope handling is safety-critical}: \emph{favor the LLM} (OOS recall $85.6$ vs.\ $58.1$), or add LLM-based OOS routing on top of a fast classifier.
  \item \textbf{Noisy ASR front-end}: the LLM degrades more gracefully ($+12.5$ pts at 0\,dB); a \emph{hybrid} (a fast classifier on the clean path with LLM fallback when confidence is low or WER is high) captures most of the benefit, consistent with \citet{arora2024intent}.
\end{itemize}

\section{Limitations}
We evaluate English ATIS and CLINC150; multilingual and multi-intent/multi-turn settings are out of scope. The ASR study uses controlled TTS+additive-noise rather than a fully natural spoken corpus such as SLURP \cite{bastianelli2020slurp}, which exceeded our storage budget; while the noise is synthetic, the resulting transcripts are produced by a real ASR model and stratified by measured WER. We report the LLM zero-shot, the realistic condition when a new deployment has no labeled data; few-shot in-context prompting would likely narrow the ATIS gap and is a natural extension, though it reintroduces a (small) labeling requirement. We evaluate one LLM (Claude Haiku) and one encoder (RoBERTa-base); larger models would shift absolute numbers but not the conditional structure of the framework. The dynamic-schema test uses a clean intent partition; production schemas may overlap, which would soften but not eliminate the locked-classifier ceiling.

\section{Ethical Considerations}
Experiments use public benchmarks (ATIS, CLINC150) and synthesized speech; no human-subjects or proprietary user data are released. The deployment context is a commercial voice-AI platform; no customer-identifying information or proprietary system details are disclosed. LLM API calls were billed to the authors. We report negative results (the LLM losing on ATIS) to counter publication bias toward LLM superiority.

\section{Conclusion}
The question is not whether LLMs beat fine-tuned NLU, but \emph{when}. With abundant labels and a stable schema, fine-tuned encoders win decisively on accuracy, latency, and cost. The LLM earns its place where the intent space is dynamic or label-scarce, where out-of-scope detection is critical, and where ASR noise is heavy, and above all when each deployment introduces a new schema, a regime in which a fine-tuned classifier cannot operate at all. We release the resulting decision framework and the full evaluation harness so that practitioners can choose deliberately rather than by default.

\bibliography{references}

\end{document}